\documentclass[runningheads]{llncs}
\usepackage{graphicx}
\usepackage{multirow}
\usepackage{booktabs,tabularx}
\newcolumntype{C}{>{\centering\arraybackslash}X}
\usepackage[font=small,skip=0pt]{caption}
\usepackage{enumitem}
\usepackage{pifont}
\usepackage{array}
\newcolumntype{P}[1]{>{\centering\arraybackslash}p{#1}}

\usepackage{subfig}
\DeclareSymbolFont{bbold}{U}{bbold}{m}{n}
\DeclareSymbolFontAlphabet{\mathbbold}{bbold}
\usepackage[title,toc,titletoc,header]{appendix}
\begin{document}
\title{Continual Class Incremental Learning for CT Thoracic Segmentation}
%
%
\author{Abdelrahman Elskhawy\inst{1,2}\and
Aneta Lisowska\inst{2}\and
Matthias Keicher\inst{1}\and
Joseph Henry\inst{2}\and
Paul Thomson\inst{2}\and
Nassir Navab\inst{1,3}}


\authorrunning{A.Elskhawy et al.}

\institute{Computer Aided Medical Procedures,  Technische Universität München \\  \email{a.elskhawy@tum.de, matthias.keicher@tum.de}\and
Canon Medical Research Europe ltd.
\email{Aneta.Lisowska@eu.medical.canon, Joseph.Henry@eu.medical.canon, and Paul.Thomson@eu.medical.canon}\and
Computer Aided Medical Procedures, Johns Hopkins University, Baltimore, USA
\email{navab@cs.tum.edu}
}

\maketitle              

\begin{abstract}

Deep learning organ segmentation approaches require large amounts of annotated training data, which is limited in supply due to reasons of confidentiality and the time required for expert manual annotation.  Therefore, being able to train models incrementally without having access to previously used data is desirable. A common form of sequential training is fine tuning (FT). In this setting, a model learns a new task effectively, but loses performance on previously learned tasks. 
The Learning without Forgetting (LwF) approach addresses this issue via replaying its own prediction for past tasks during model training. In this work, we evaluate FT and LwF for class incremental learning in multi-organ segmentation using the publicly available AAPM dataset. We show that LwF can successfully retain knowledge on previous segmentations, however, its ability to learn a new class decreases with the addition of each class. To address this problem we propose an adversarial continual learning segmentation approach (ACLSeg), which disentangles feature space into task-specific and task-invariant features. This enables preservation of performance on past tasks and effective acquisition of new knowledge.

\end{abstract}
\section{Introduction}
\begin{figure}
\includegraphics[width=\textwidth]{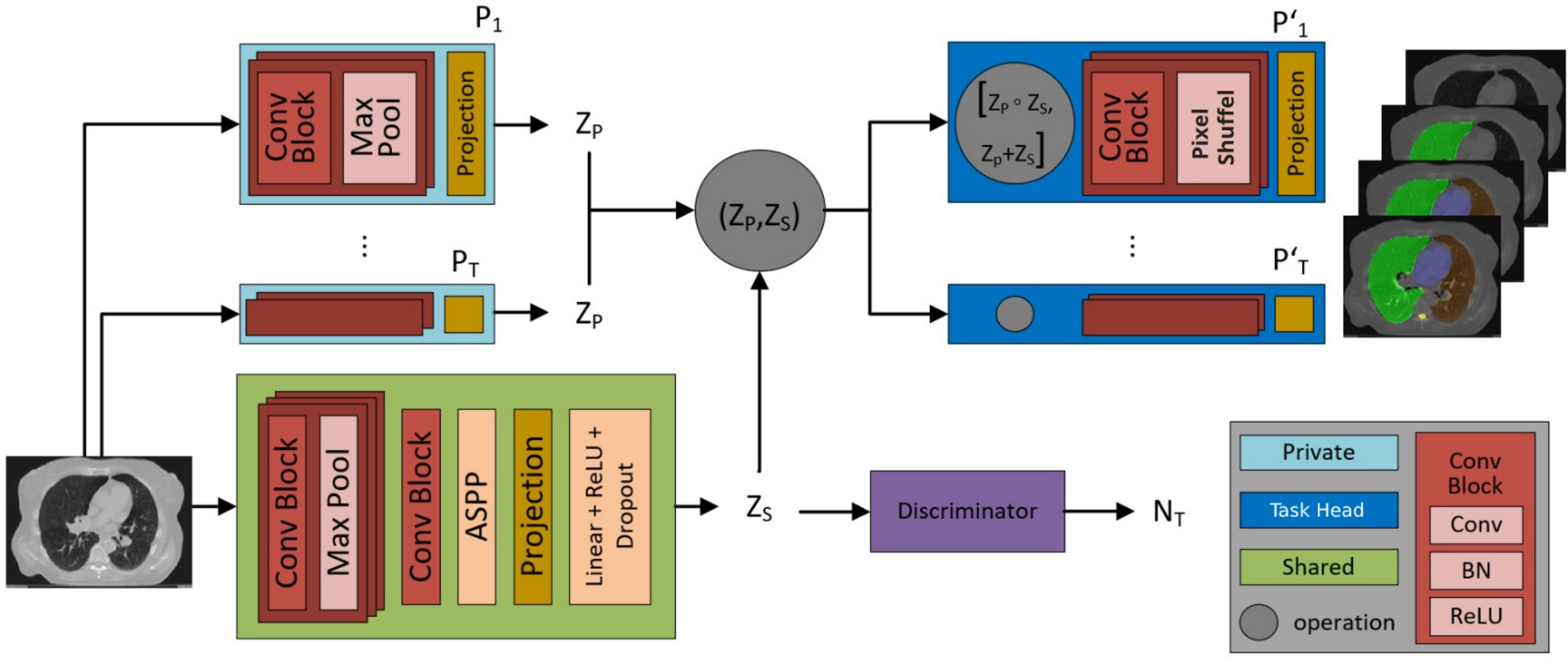}
\caption{ACLSeg architecture. Considering a sequence of $T$ tasks, a private module ($P$) and a task head ($P'$) are added for each task, while the shared module ($S$) is common between all tasks. Both $S$ and the $P$ for each task receive the input CT slice and generate the shared ($Z_S$) and private ($Z_P$) embeddings respectively. For each task head, $Z_S$ and $Z_P$ are multiplied and added element-wise, then concatenated to be further processed using Conv block and PixelShuffle modules to generate the final output. The discriminator receives $Z_S$ only and tries to predict the task label $N_T$ in an adversarial $minmax$ game with the shared module. A projection layer towards the end of each module reduces the number of channels to 1. } \label{fig1}

\end{figure}

The best performing deep learning solutions are trained on a large number of annotated training examples. However, it might be infeasible to annotate large amounts of data to train specialised models from scratch for each new medical imaging problem.  This can be due to privacy regulations that impose constraints on sharing patients' sensitive data, fragmented healthcare systems, and/or the time and expense required for expert manual annotation. Therefore, being able to train models incrementally without having access to previously used data is desirable.

The most common form of sequential training is fine tuning (FT). In this setting, a pre-trained model can learn a new task effectively from a smaller amount of data, but at the cost of losing its ability to perform previously learned tasks; a phenomenon known as catastrophic forgetting \cite{mccloskey1989catastrophic}. 
Continual learning (CL) approaches intend to address this issue either via structural growth such as \cite{rusu2016progressive,yoon2017lifelong}, which relies on adding task-specific modules, regularisation-based methods such as \cite{kirkpatrick2017overcoming,baweja2018towards}, which penalise significant changes to the previous tasks' representations, or replay-based methods such as \cite{rebuffi2017icarl,shin2017continual} which replay previous data either explicitly or via pseudo-rehearsal. For the purposes of this work, we do not discuss explicit replay-based methods as we assume that direct access to previous data is not possible.
Learning without Forgetting (LwF) \cite{li2017learning}, which combines both regularisation-based and pseudo-rehearsal techniques, has been the state-of-the-art medical imaging continual learning method for situations in which access to previous training data is not possible. It has shown promising results in medical imaging for both incremental domain learning \cite{lenga2020continual} and incremental class learning \cite{ozdemir2019extending}. However, it has not been previously evaluated on the incremental class learning problem with a task sequence exceeding two tasks. 

\noindent In this work we:

\begin{itemize}[label=$\bullet$]
    \item Evaluate LwF on a sequence of 5 segmentation tasks and show that it struggles to accommodate more information as the number of tasks increases. 
    \item Adopt the Adversarial Continual Learning (ACL) \cite{ebrahimi2020adversarial} approach  to work for segmentation problems and call it ACL Segmentation (ACLSeg)
    \item Compare the ACLSeg approach with FT and LwF and demonstrate that ACLSeg retains previously learned knowledge while being able to learn newly added tasks. %
    \item Explore task-order robustness for both ACLSeg and LwF. 
    \ 
\end{itemize}

\section{Related work}
\subsubsection{Continual Learning in the medical domain.} Although there are various CL approaches proposed for natural image classification tasks \cite{parisi2019continual}, not many of them have been applied to medical image segmentation. 
For domain incremental learning,  Ozgung et al \cite{ozgun2020importance} proposed learning rate regularisation to Memory Aware Synapses \cite{aljundi2018memory} to perform MRI brain segmentation. 
Lenga et al have shown that LwF outperforms elastic weight consolidation (EwC) \cite{kirkpatrick2017overcoming},  when applied to incremental X-ray domain learning \cite{lenga2020continual}. 

In the incremental class learning setting, Baweja et al \cite{baweja2018towards} used EWC to sequentially learn cerebrospinal fluid segmentation followed by grey and white matter segmentation tasks. 
Ozdemir and Goksel  \cite{ozdemir2019extending} applied Learning without Forgetting (LwF) to sequential learning of tibia and femur bone in MRI of the knee. The authors suggested that LwF is a viable CL solution when sharing patient data is not possible due to privacy concerns. When retention of representative samples from past datasets is possible, the authors suggested to use AeiSeg, which extends LwF via sample replay, leading to improved knowledge preservation. In this work we assume that there is no access to previous data, therefore we do not include AeiSeg in our comparison. 
\subsubsection{Latent space disentanglement.} Multi-view learning \cite{li2018survey} exploits different modalities of the data to maximise the performance. For class incremental learning, factorising the data representation into both shared and task-specific parts helps prevent forgetting. While the learned shared representation is less susceptible to forgetting, as it is task-invariant, preventing forgetting in the private representations can be achieved by using small sub-modules per class that are frozen upon finishing learning that specific class. Latent space factorisation can be achieved by either Adversarial training as in \cite{ebrahimi2020adversarial}, orthogonality constraints as in \cite{salzmann2010factorized}, or both to ensure complete enforced factorisation.

\section{ACL Segmentation (ACLSeg)} \label{aclseg}

Our objective is to learn to segment $T$ organs in CT scans in a sequential manner. To achieve this, we build upon the ACL approach \cite{ebrahimi2020adversarial}, initially developed for incremental classification problems on MNiST \cite{lecun2010mnist} and CIFAR\cite{Krizhevsky09learningmultiple},  and adopt it to solve segmentation problems. Fig.~\ref{fig1} shows the architecture for ACLSeg including the modifications that are described in this section.    \\

Consider a sequence of $T$ tasks to be learned one task at a time. For the very first task, the model consists of the shared module, the discriminator, one private module $P_1$, and one task head $P'_1$. The discriminator attempts to predict the task label in a $minmax$ game with the shared module. When adding new segmentation tasks we add a task head and a private module,  while the shared module remains common to all tasks.

The main idea of ACL is to learn a disjoint latent space representation composed of task-invariant (shared) latent space, represented by $Z_S$, and task-specific (private) latent space, represented by $Z_P$. A task-specific head receives both $Z_S$ and $Z_P$ to generate the final output. The objective function for ACL is:
\begin{equation}
\mathcal{L}_{\mathrm{ACLSeg}}=\lambda_{1} \mathcal{L}_{\mathrm{task}}+\lambda_{2} \mathcal{L}_{\mathrm{adv}}+\lambda_{3} \mathcal{L}_{\mathrm{diff}}
\end{equation}

Where $\mathcal{L}_{\mathrm{task}}$ is the task loss (Binary Cross Entropy loss is used for each segmentation task), $\mathcal{L}_{\mathrm{adv}}$ is the $T$-way classification cross-entropy adversarial loss, and $\mathcal{L}_{\mathrm{diff}}$ is an orthogonality constraint introduced in \cite{salzmann2010factorized}, also known as the difference loss \cite{bousmalis2016domain} in domain adaptation literature.  $\mathcal{L}_{\mathrm{diff}}$ ensures further factorisation of the shared and private features. $\mathcal{L}_{\mathrm{adv}}$ and $\mathcal{L}_{\mathrm{diff}}$ are described in detail in the supplementary materials and \cite{ebrahimi2020adversarial}. $\lambda_{1}$, $\lambda_{2}$, and $\lambda_{3}$ are regularisers to control the strength of each loss component. 

\subsubsection{Adaptation to CT segmentation}
To adapt ACL to segmentation of CT slices, we introduce a few changes to the different components of the original architecture, taking into consideration the final model size.  
\\ \\
\textit{Shared Module} To enrich the extracted features in the shared module, we used an encoder-based module with Atrous Spatial Pyramid Pooling (ASPP) \cite{chen2018encoder}. ASPP allows explicit control over the resolution of extracted features and adjusts a filter's field of view to capture multi-context information at reduced computational cost. This is desirable due to the size of medical data, and the need to segment small anatomical structures. \\

\textit{Task Heads} We introduce two modifications to the task heads: \textbf{A)} While the original ACL paper proposes concatenating $Z_S$ and $Z_P$ to generate the final output, \cite{jayakumar2020multiplicative} suggested that this is not the optimal way to fuse multiple streams of information. In order to ensure enriched representations in each task head, we replace the concatenation operation with additive and multiplicative counter parts, i.e. $Z_S$ and $Z_P$ are multiplied and added, then concatenated over the channel dimension and tehn passed to the respective task head. Given the two vectors $Z_S$ and $Z_P$ of length $Latent\_dim$ each, the input to the respective task head is the concatenation of both $(Z_S \odot Z_P)$ and $(Z_S \oplus Z_P)$ along the channel dimension resulting in a two-channel feature map of final size $Latent\_dim \times 2$. \textbf{B)} We designed the task head to be compact to make the model scalable as we  add a task head for each newly added task. To achieve this, we upsample the private and shared embeddings in two stages with a 4X upsampling factor at each stage. Unlike the U-Net architecture \cite{ronneberger2015u}, where multi-scale features are fused via skip connections from the encoder to the decoder, our model architecture relies on the output from the embedding vectors only. To improve over the upsampling and to be able to recover some of the segmentation details with such a large upsampling factor, we replace the Convolution Transpose upsampling with sub-pixel convolutions proposed in \cite{shi2016real}.  This approach uses regular convolution layers followed by a Phase Shift reshaping operation which solves the checkerboard artifacts and improves the segmentation results as shown in \cite{chen2018semantic,gao2017pixel,lachinov2019segmentation}. 
For further details on how each of these modifications contributed to the segmentation performance please refer to table 1 in the Supplementary Material.

\subsection{Evaluation metrics}
In order to assess a CL system, the system needs to be evaluated on two different aspects. First, segmentation quality, which is useful for tracking the running segmentation score to ensure that the model is providing meaningful segmentation. We report this as the Dice Coefficient (DC). Second, knowledge retention and the ability to learn new information. For this, we adopt the metrics proposed in \cite{kemker2018measuring} with slight modifications to $\Omega_{n e w}$ calculations, in which we normalise the value to fall in the range $[0,1]$.  This makes the three $\Omega$ values comparable across different continual learning approaches.  Therefore, our modified knowledge retention metrics are:

\begin{equation}
\Omega_{b a s e}=\frac{1}{T-1} \sum_{i=2}^{T} \frac{\alpha_{b a s e, i}}{\alpha_{i d e a l, base}} \label{eq:omega_base}   
\end{equation}

\begin{equation}
\Omega_{n e w}=\frac{1}{T-1} \sum_{i=2}^{T} \frac{\alpha_{new, i}}{\alpha_{i d e a l,i}}        \label{eq:omega_new}
\end{equation}

\begin{equation}
\Omega_{a l l}=\frac{1}{T-1} \sum_{i=2}^{T} \frac{\overline{\alpha_{all, 0:i}}}{\overline{\alpha_{ideal, 0:i}}}
\label{eq:omega_all}
\end{equation} \medskip

where $T$ is the total number of classes, $\alpha_{base, i}$ is the DC of the first class after $i$ classes have been learned, $\alpha_{new, i}$ is the DC of class $i$ immediately after it is learned,  $\overline{\alpha_{all, 0:i}}$ is the mean DC of all the classes that have been seen so far up to and including step $i$, $\overline{\alpha_{ideal, 0:i}}$ is the offline mean DC of all the classes that have been seen so far up to and including step i, by jointly training the model on all the available data at once, and $\alpha_{ideal, base}$, and $\alpha_{ideal, i}$ are the offline ideal DC of the base and the $i^{th}$ class respectively.

$\Omega_{base}$ measures the model retention of the first learned class after learning subsequent classes, $\Omega_{new}$ measures the model ability to learn new classes, and 
$\Omega_{all}$ computes how well a model can both retain prior knowledge and acquire new information. All $\Omega$ values $\in[0,1]$ unless a CL model exceeds the upper bound. Since the $\alpha_{ideal, n}$ is obtained from offline training the same model on all the data at once, the architectural choice of the model does not affect our comparison.  

\section{Experiments and Results}
\subsection{Experimental Setup}
\subsubsection{Datasets} 
We experiment with the publicly available AAPM Thoracic auto-segmentation challenge dataset (AAPM) \cite{yang2017data}. The AAPM dataset has segmentations for 5 organs: spinal cord, right lung, left lung, heart, and oesophagus. The training set is composed of 30 scans, which are further split into 5 subsets, one for each class, 6 validation and 24 testing CT scans.\\ \\
\textbf{Training scheme and hyperparameters} For all experiments, the models were trained to convergence using EarlyStopping on a validation dataset. We chose an initial learning rate (lr) of 1e-3, which is reduced with a factor of 3 on validation loss plateau. The algorithm is implemented using Pytorch, and we train the network using the Adam optimiser \cite{kingma2014adam}. The inputs were normalised, and resized to 256x256 instead of 512x512 with no other data augmentation techniques applied. $Latent\_dim$ is chosen to be 256, and $\lambda_{1}$, $\lambda_{2}$, and $\lambda_{3}$ are 1, 0.05, and 0.3 respectively. Empirically, higher values of $\lambda_{2}$ would render the adversarial training unstable, while higher values of $\lambda_{3}$ would concentrate most of the information in private modules leading to sub-optimal latent space separation. We average all the demonstrated results across three runs to report the mean and standard deviation.

\subsubsection{Baselines} 
We compare ACLSeg with LwF which represents the state-of-the-art in class incremental learning for segmentation problems\cite{ozdemir2019extending} when access to previous data is not possible. For this purpose, we adopted the Multi-head U-Net structure proposed in \cite{ozdemir2019extending}. We used Binary Cross Entropy as the segmentation loss for each task, and regressed the class probabilities (logits) of the previous model using mean square loss as our knowledge distillation loss, as proposed in \cite{shmelkov2017incremental}. We also perform naive Fine Tuning (FT) in which a single model is trained sequentially with no forgetting prevention techniques which serves as a lower bound. For the upper bound (ideal), we jointly train the model in a multitask setting with all the available data.  
\begin{center}
\begin{table}[t]
\caption{$\Omega$ scores, (Std. dev. of 3 runs), and overall dice score of the final model for class incremental learning on 5 classes}\label{tab_cli}
\begin{tabularx}{\linewidth}{ c *{5}{C} }
    \toprule
    & $\Omega_{base}$ & $\Omega_{new}$ & $\Omega_{all}$ & Overall dice score \\ 
    \midrule
FT   &   0.03(0.000)  &      0.93(0.090) & 0.33(0.020) & 0.14(0.008)      \\
LwF  &  \textbf{1.09(0.010)} &  0.82(0.005) &   0.96(0.026) & 0.75(0.010) \\
ACLSeg   &  1.00(0.005)&   \textbf{0.96(0.006)} &   \textbf{0.99(0.004)}& \textbf{0.80(0.005)}     \\
    \bottomrule
\end{tabularx}
\end{table}
\end{center}
\subsection{Results}
\begin{figure}
\includegraphics[width=\textwidth]{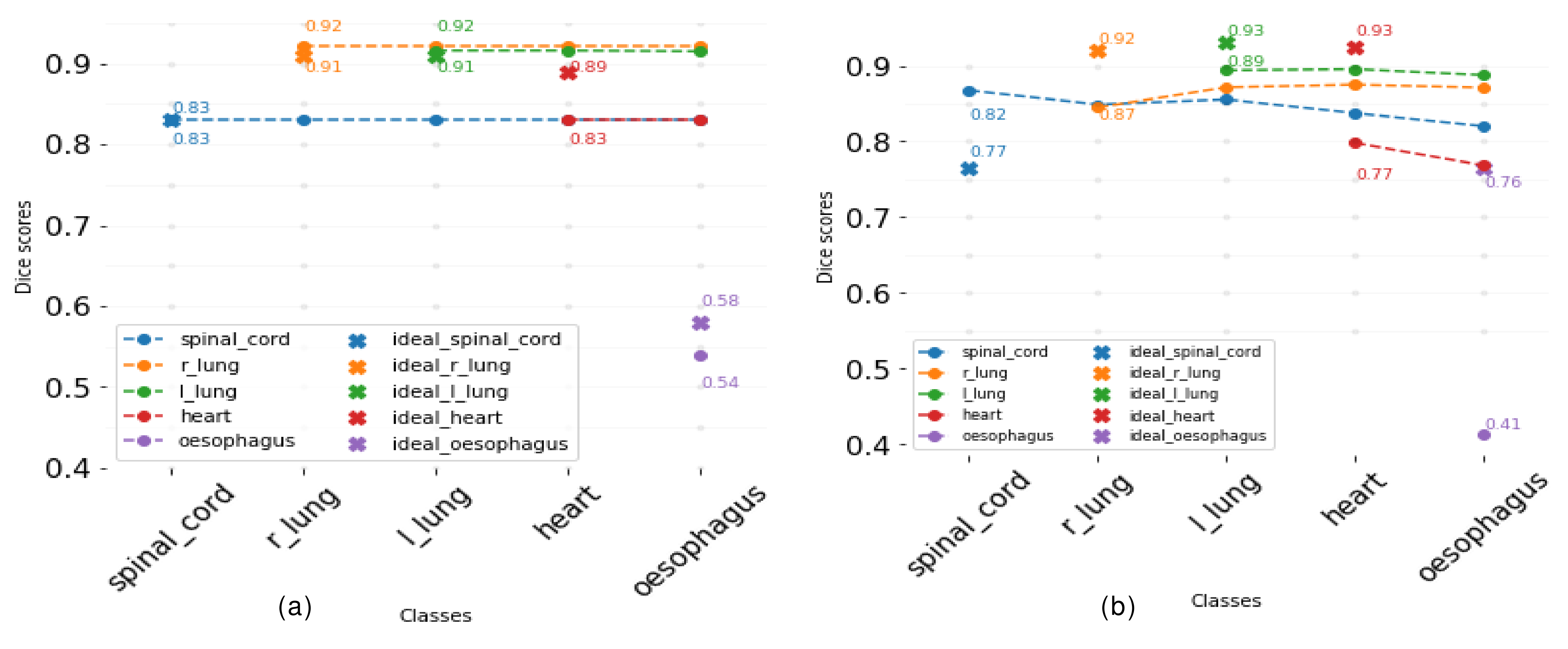}
\caption{Dice coefficients for a model trained sequentially on the AAPM dataset along with the corresponding ideal dice coefficients obtained from offline training.(a)  ACLSeg, (b) LwF } \label{fig_dc}
\end{figure}

\begin{figure}[hbt]
    \centering
    \includegraphics[width=0.99\textwidth,trim={0.8cm 0.0cm 1.5cm 0.0cm}, clip]{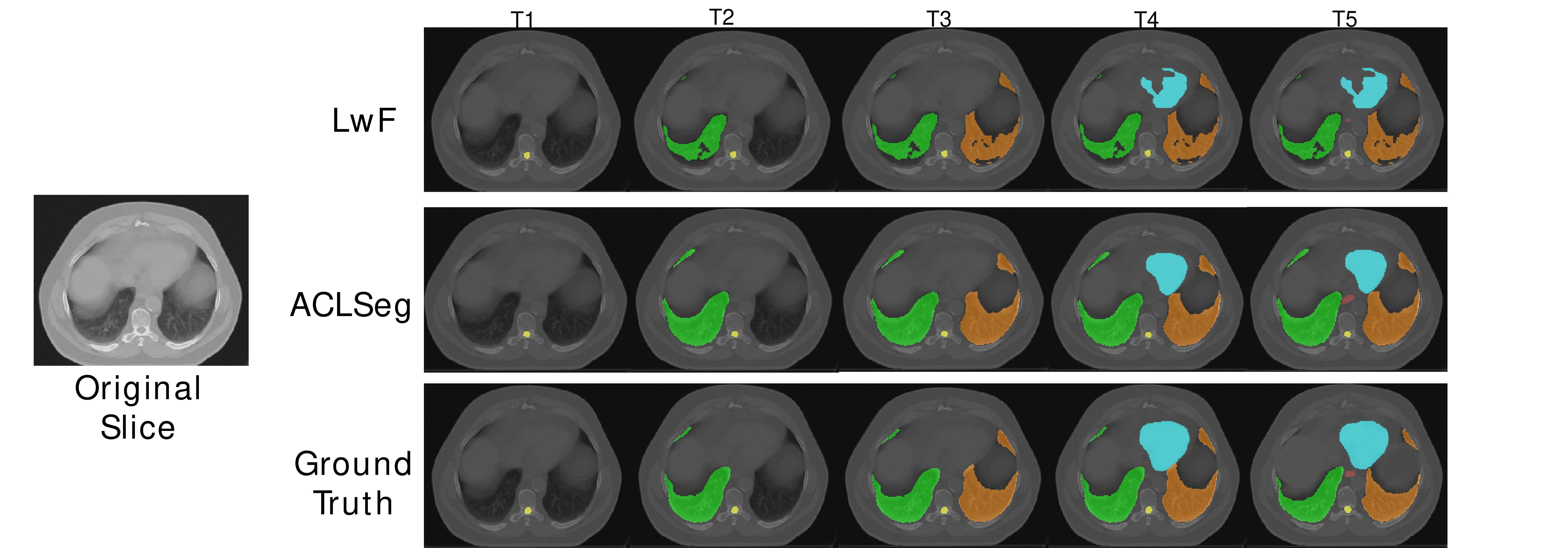}
    \caption{Ground truth and segmentation results for a given input slice using LwF and ACLSeg after learning each task in OrderA}
    \label{figseg}
\end{figure} 
\subsubsection{5-Split AAPM } 
\begin{figure}

\includegraphics[width=0.95\textwidth,trim={0.8cm 0.0cm 1.5cm 0.0cm}, clip]{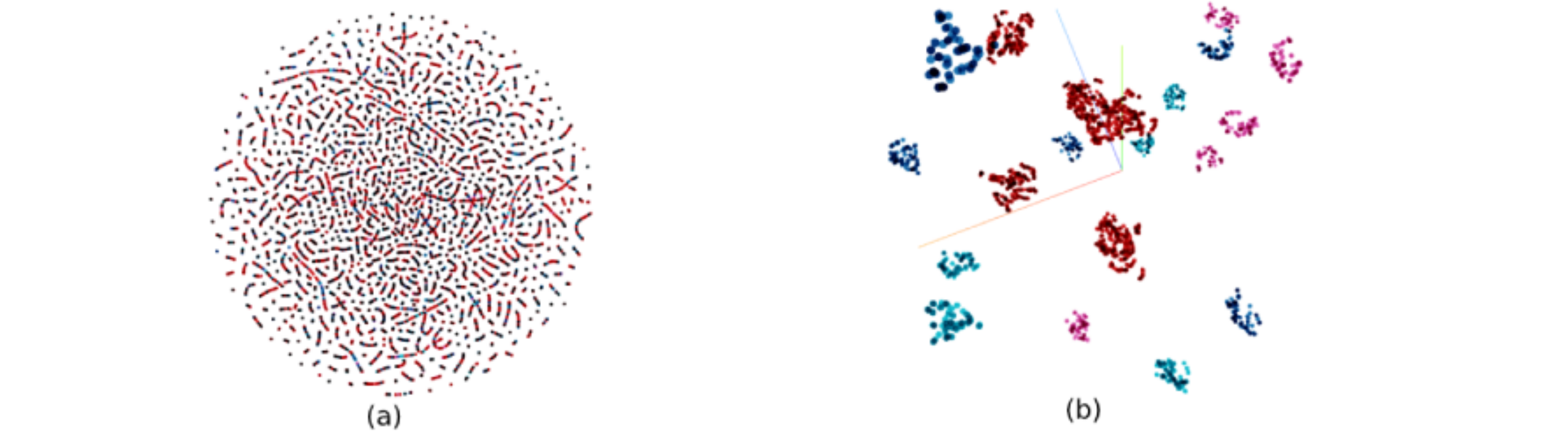}
\caption{T-sne visualisation of the embeddings generated by a) the shared module and b) private modules} \label{figtsne}
\end{figure}

We split the training dataset into 5 subsets, one for each of the classes, and learned one class at a time. Table \ref{tab_cli} shows the obtained Omega scores along with the overall dice score of the final model. The results are reported on the sequence: Spinal Cord, Right Lung, Left Lung, Heart, and Oesophagus (OrderA). We observe that although LwF is able to retain the performance of the base task, shown by the large $\Omega_{base}$ score, it struggles to accommodate new information as more classes are added. Fig. \ref{fig_dc} shows the change in dice scores after the addition of new classes. We observe that LwF performance on some of the previous classes exhibits a small degradation with the addition of new classes, and does not fully learn the $4^{th}$ and $5^{th}$ classes (See Figure \ref{figseg}). In contrast, the FT approach shows a high $\Omega_{new}$ score as it focuses on learning the new classes, at the expense of losing previous information. However $\Omega_{new}$ score does not reach the ideal score, possibly due to the differences between subsequent tasks, which gives unfavorable initialisation of the model weights for learning the new task. ACLSeg combines both capabilities by being able to retain a consistent performance on all the previously learned tasks, while having the ability to reach near-ideal performance on subsequent tasks as shown in Fig. \ref{fig_dc}, and reflected in the 5\% increase in the overall dice score of the final model compared to LwF in Table \ref{tab_cli}. Our interpretation is that this might be due to the disentanglement of the latent space which preserves the task-related knowledge in the respective private module, while being able to update the shared module with only the task-invariant information, hence preventing catastrophic forgetting.
Fig. \ref{figtsne} shows the t-SNE visualisation of the generated embeddings, which shows that the shared embeddings form a uniform distribution of samples belonging to all classes which can not be uncovered, while the private modules are successful in uncovering class labels in their latent space. We point out that although ACLSeg shows zero forgetting, it struggles to learn the last class (Oesophagus) due to its complexity and severe under-representation in the dataset. This is also true for LwF and ideal training and leaves room for improvement.  


\subsubsection{Task-order robustness} Yoon et al showed that CL model performance significantly varies based on the order in which the tasks are learned  \cite{yoon2019scalable}. Since this large variance might cause an issue in the medical domain, we investigate different task orders. We pick two different sequences in addition to OrderA, and report our results in Table \ref{task_order}. OrderB represents the sequence "Oesophagus, Heart, Left Lung, Right Lung, and Spinal Cord" which starts with the hardest-to-segment class. OrderC represents the sequence "Left Lung,  Right Lung, Spinal Cord, Heart, and Oesophagus" which starts with an easy-to-segment class followed by a medium difficulty one, while OrderA starts with a medium difficulty class followed by an easy one. From Table \ref{task_order} we observe that starting with a hard-to-segment task has an effect on the base score as the model was not able to fully learn the base class, Oesophagus in this case, however, it was able to maintain the performance on this class till the end of sequential training (see supplementary material for detailed results).

\begin{center}
\begin{table}[t]
\caption{$\Omega$ scores and (Std. dev. of 3 runs) for different task orders}\label{task_order}
\begin{tabularx}{\linewidth}{P{1.3cm}P{1.7cm}P{1.8cm}P{1.7cm}P{1.7cm}P{1.7cm}P{1.7cm}} 
    \toprule
    & \multicolumn{2}{c}{$\Omega_{base}$}
            & \multicolumn{2}{c}{$\Omega_{new}$}
                    & \multicolumn{2}{c}{$\Omega_{all}$} \\ 
    \cmidrule(l ){2 -3} \cmidrule(l){4-5} \cmidrule(l){6-7}
    & ACLSeg & LwF & ACLSeg & LwF & ACLSeg & LwF \\           
    \midrule
OrderA   &  1.01(0.005) & \textbf{1.09(0.010)} &  \textbf{0.96(0.006)}       &  0.82(0.005)  & \textbf{0.99(0.004)}       & 0.96(0.026) \\

OrderB   &   \textbf{0.82(0.010)} & 0.53(0.140) & \textbf{0.98(0.001)}           &  0.83(0.080)  & \textbf{0.93(0.004)}     & 0.70(0.040)  \\

OrderC   &   0.99(0.002)  & \textbf{1.0(0.004)} &  \textbf{0.94(0.005)} & 0.84(0.020) &  \textbf{0.99(0.003)}& 0.94(0.009) \\
    \bottomrule
\end{tabularx}
\end{table}
\end{center}

\section{Conclusion}
We adapted an adversarial continual learning approach to medical data (ACLSeg) and evaluated it on an incremental thoracic segmentation problem. We demonstrated that ACLSeg retains knowledge equally as well as LwF, while being able to achieve better performance on newly added tasks. For both approaches task order affects the anatomy segmentation performance, however for ACLSeg the knowledge retention is preserved.
We also showed that ACLSeg has a disentangled latent space that is composed of task-invariant and task-specific representations which might be useful for model explainability and privacy preservation.
%
%
%
\bibliographystyle{splncs04}
\bibliography{ref.bib}
%
\newpage

\title{Supplementary Material: Continual Class Incremental Learning for CT Thoracic Segmentation} \titlerunning{SM: Continual Class Incremental Learning for CT Thoracic Segmentation }
\author{Abdelrahman Elskhawy\inst{1,2}\and
Aneta Lisowska\inst{2}\and
Matthias Keicher\inst{1}\and
Joseph Henry\inst{2}\and
Paul Thomson\inst{2}\and
Nassir Navab\inst{1,3}}

\authorrunning{A.Elskhawy et al.}
%
\institute{Computer Aided Medical Procedures,  Technische Universität München \and
Canon Medical Research Europe ltd. \and
Computer Aided Medical Procedures, Johns Hopkins University, Baltimore, USA
}

\maketitle
\section{ACLSeg loss functions}
Consider a sequence of $T$ tasks to be learned sequentially. Each task's data $D_t$ is represented by the tuple $\mathcal{D}_{t}=\left\{\left(\mathbf{X}_{i}^{k}, \mathbf{Y}_{i}^{k}, \mathbf{T}_{i}^{k}\right)_{i=1}^{n_{k}}\right\}$ where $K$ is the task number, $n$ is the number of input samples $\left(\mathbf{X}^{k} \in \mathcal{X}\right)$, output labels $\left(\mathbf{Y}^{k} \in \mathcal{Y}\right)$, and task label $\left(\mathbf{T}^{k} \in \mathcal{T}\right)$. The objective is to learn a mapping $f_{\theta}: \mathcal{X} \rightarrow \mathcal{Y}$ for each task to map the input to its target output. To learn $f_{\theta}$, we use the Binary Cross Entropy (BCE) loss which is defined as


\begin{equation}
    \mathcal{L}_{{task }}=- (1/n) \sum_{i=1}^{n} \log (\sigma(f_{\theta}^{k}(x_{i}^{k})))
\end{equation}
where $\sigma$ is the sigmoid function. 

In order to disentangle the latent space into $Z_S$ and $Z_P$, we learn the shared mapping $\left(S_{\theta_{S}}: \mathcal{X} \rightarrow \mathbf{Z}_{S}\right)$ and the private mapping $\left(P_{\theta_{S}}: \mathcal{X} \rightarrow \mathbf{Z}_{P}\right)$ respectively. The shared mapping is trained to generate embeddings that fool an adversarial discriminator. The discriminator $\left(D_{\theta_{D}}: \mathbf{z}_{S} \rightarrow \mathcal{T}\right)$ on the other hand tries to classify the embeddings by their task labels $\left(\mathbf{T}^{k \in\{0, \cdots, T\}}\right)$. This is achieved through the minmax game between $D$ and $S$ characterized by the cross-entropy adversarial loss described as 

\begin{equation}
    \mathcal{L}_{{adv}}=\min_{S} \max_{D} \sum_{k=0}^{T} \mathbbold{1}_{[k=t^{k}]} \log (D(S(x^{k})))
\end{equation}

where task label zero is paired with randomly generated noise features. 

To further factorize $Z_S$ and $Z_P$, the difference loss is used to prevent the shared features from appearing in the private embeddings. The diff loss is described as 

\begin{equation}
    \mathcal{L}_{\mathrm{diff}}=\sum_{k=1}^{T}\left\|\left(S\left(x^{k}\right)\right)^{\mathrm{T}} P^{k}\left(x^{k}\right)\right\|_{F}^{2}
\end{equation}

This renders the objective function for ACLSeg as described in section 3 of the paper.

\begin{equation}
\mathcal{L}_{\mathrm{ACLSeg}}=\lambda_{1} \mathcal{L}_{\mathrm{task}}+\lambda_{2} \mathcal{L}_{\mathrm{adv}}+\lambda_{3} \mathcal{L}_{\mathrm{diff}}
\end{equation}

\section{Ablation Study} 
\begin{figure}
    \centering
    \subfloat[]{{\includegraphics[width=5cm]{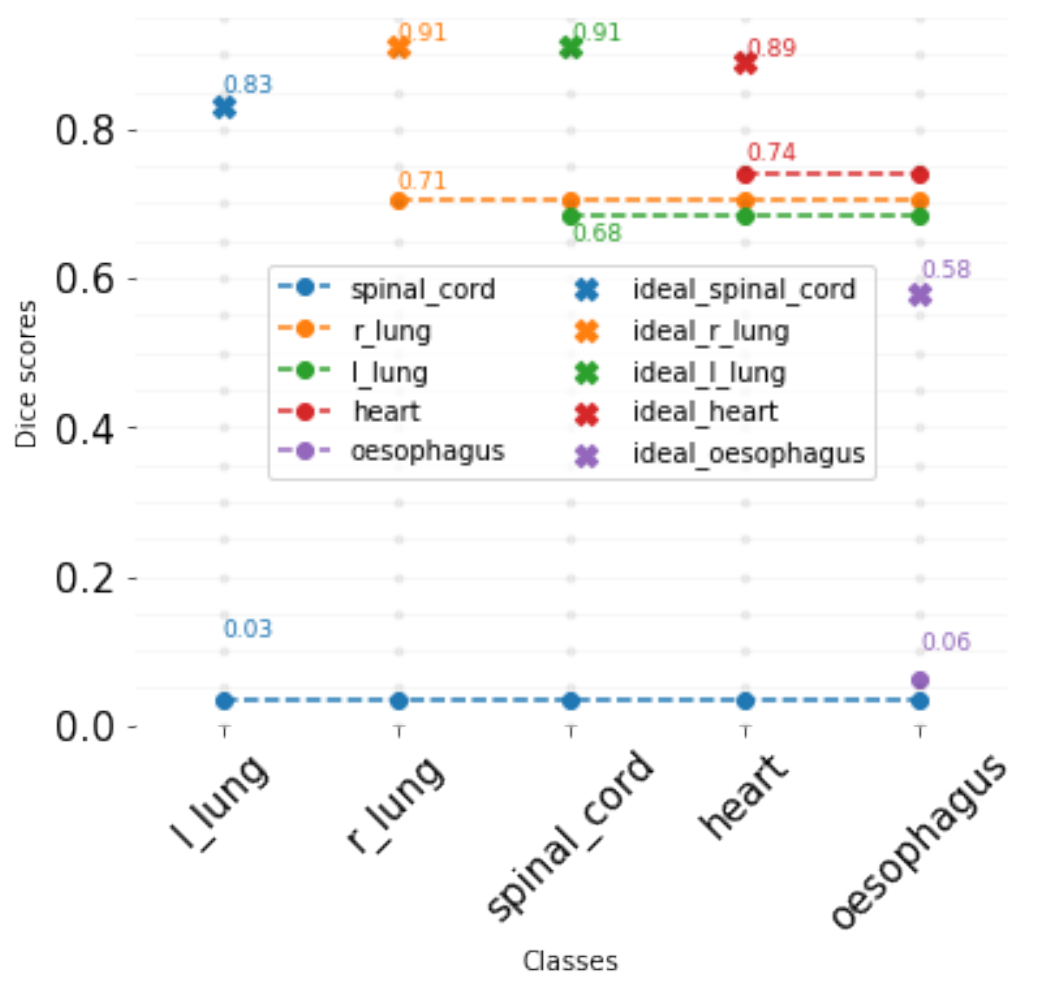} }}%
    \qquad
    \subfloat[ ]{{\includegraphics[width=5cm]{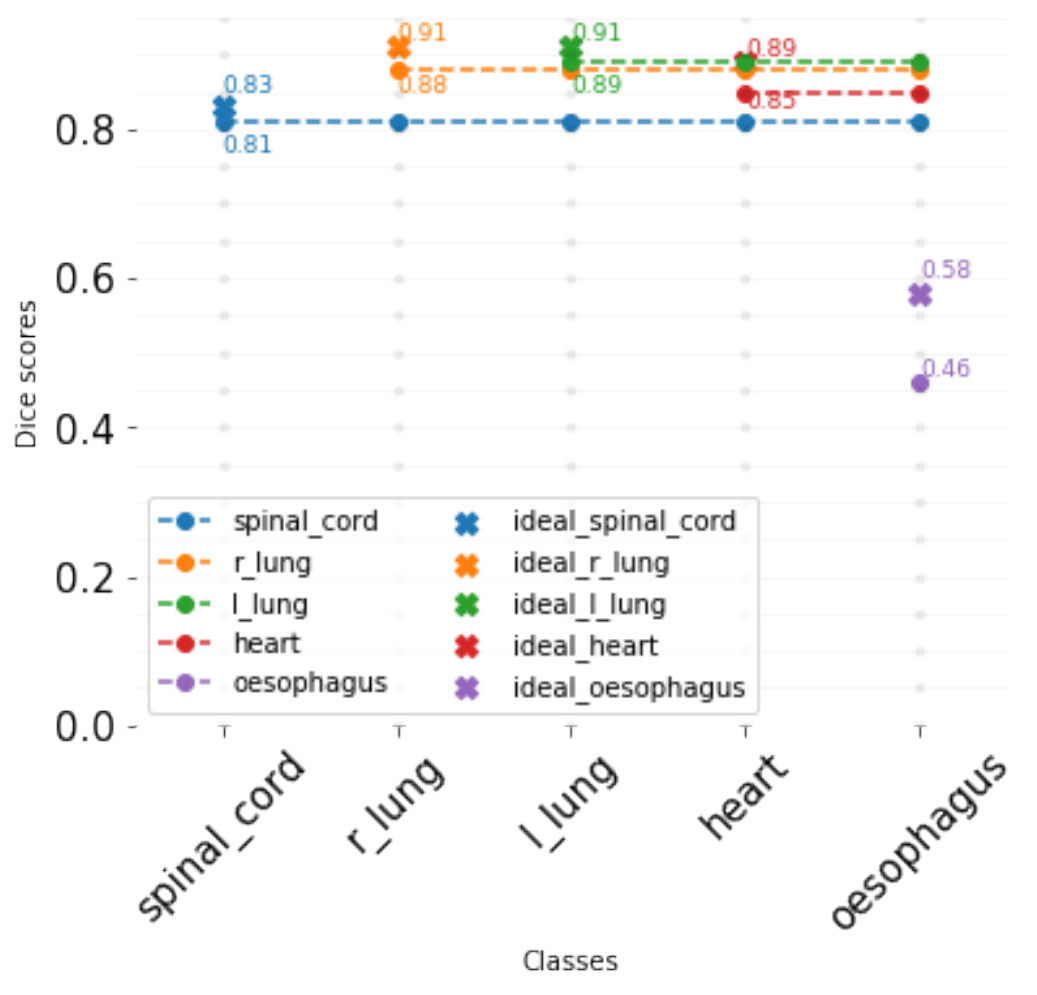} }}%
    \caption{Dice Scores for a) "Basic Enc." and b) "With ASPP \& PS"}%
    \label{fig:ablation}%
\end{figure}

In this section we show the contribution of each of the proposed modifications in section 2 in the paper. We start from replacing the MNiST feature extractor with a basic encoder (Basic Enc.) suitable for complex medical data, then we add the ASPP module and PixelShuffle (With ASPP \& PS), and finally we replace the concatenation of embeddings in task head with both the addition and multiplication (ACLSeg) which represents the final state of the model. Table \ref{tab_ablation} shows the omega scores achieved at each step. We note that even with the basic encoder structure, ACLSeg demonstrates the same information-preserving behaviour, i.e. it retains the performance it achieves across all training phases, and all the proposed modifications were introduced to obtain acceptable segmentation quality. This is also demonstrated by the dice score plots in Fig. \ref{fig:ablation}.
\begin{center}
\begin{table}[t]
\caption{Ablation study of ACLSeg on AAPM dataset}\label{tab_ablation}
\begin{tabularx}{\linewidth}{ c *{5}{C} }
    \toprule
    & $\Omega_{base}$ & $\Omega_{new}$ & $\Omega_{all}$ \\ 
    \midrule
Basic Enc.   &   0.04  &      0.57 & 0.53      \\
With ASPP \& PS  &  0.97 &  0.91 &   0.95  \\
ACLSeg   &  1.01&   0.96 &   0.99     \\
    \bottomrule
\end{tabularx}
\end{table}
\end{center}

\begin{center}
\begin{table}
\centering
\caption{Overall dice score of the final model for class incremental learning on 5 classes for different task orders}\label{tab_avg_dice}
\begin{tabularx}{8cm}{ c *{2}{C} }
    \toprule
    & \multicolumn{2}{c}{Overall Dice score} \\ 
    \cmidrule(l ){2 -3}
     & ACLSeg & LwF \\
    \midrule
Offline (Upper bound)  &  0.82(0.006) & 0.85(0.050)  \\
OrderA  &  0.80(0.005) &  0.75(0.014)   \\
OrderB  &  0.79(0.006) & 0.74(0.086) \\  
OrderC   &  0.79(0.003)  &   0.74(0.011) \\
    \bottomrule
\end{tabularx}
\end{table}
\end{center}

\begin{center}
\begin{figure}%
    \centering
    \subfloat[]{{\includegraphics[width=5cm]{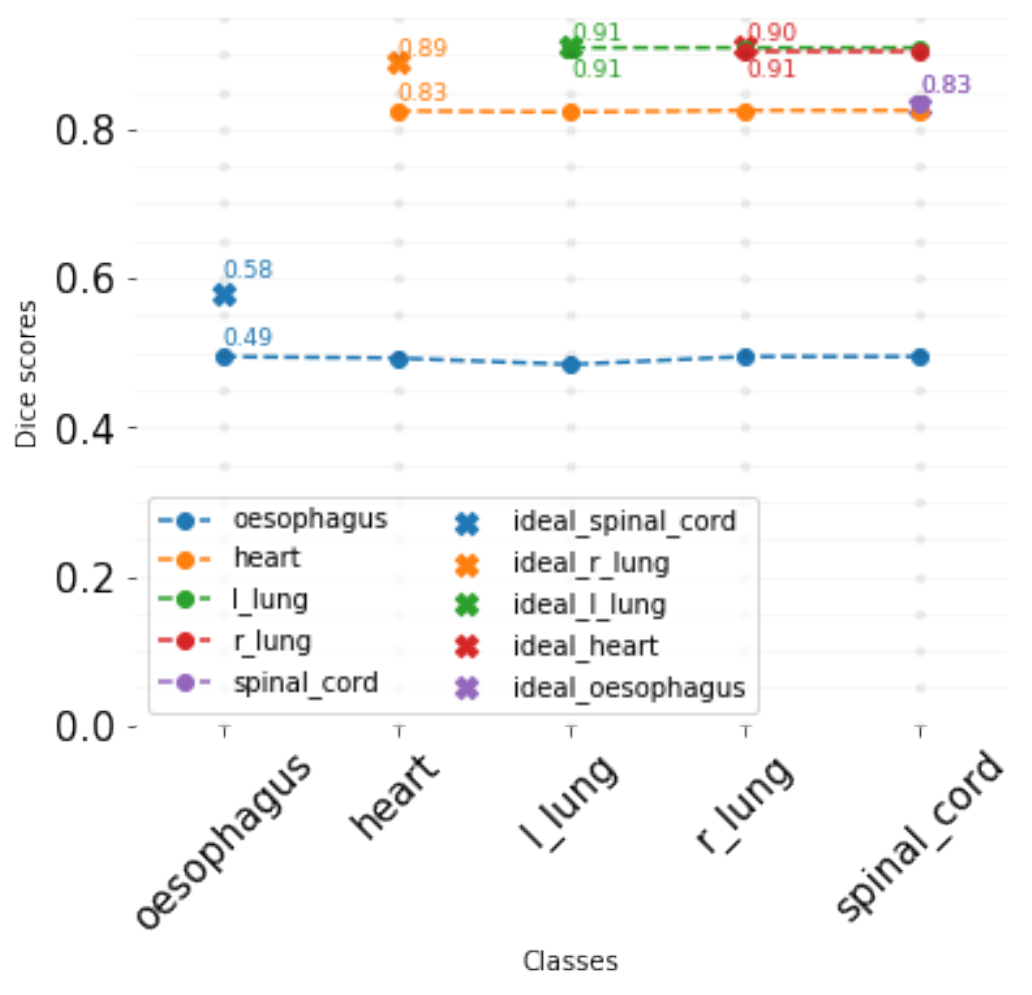} }}%
    \qquad
    \subfloat[]{{\includegraphics[width=5cm]{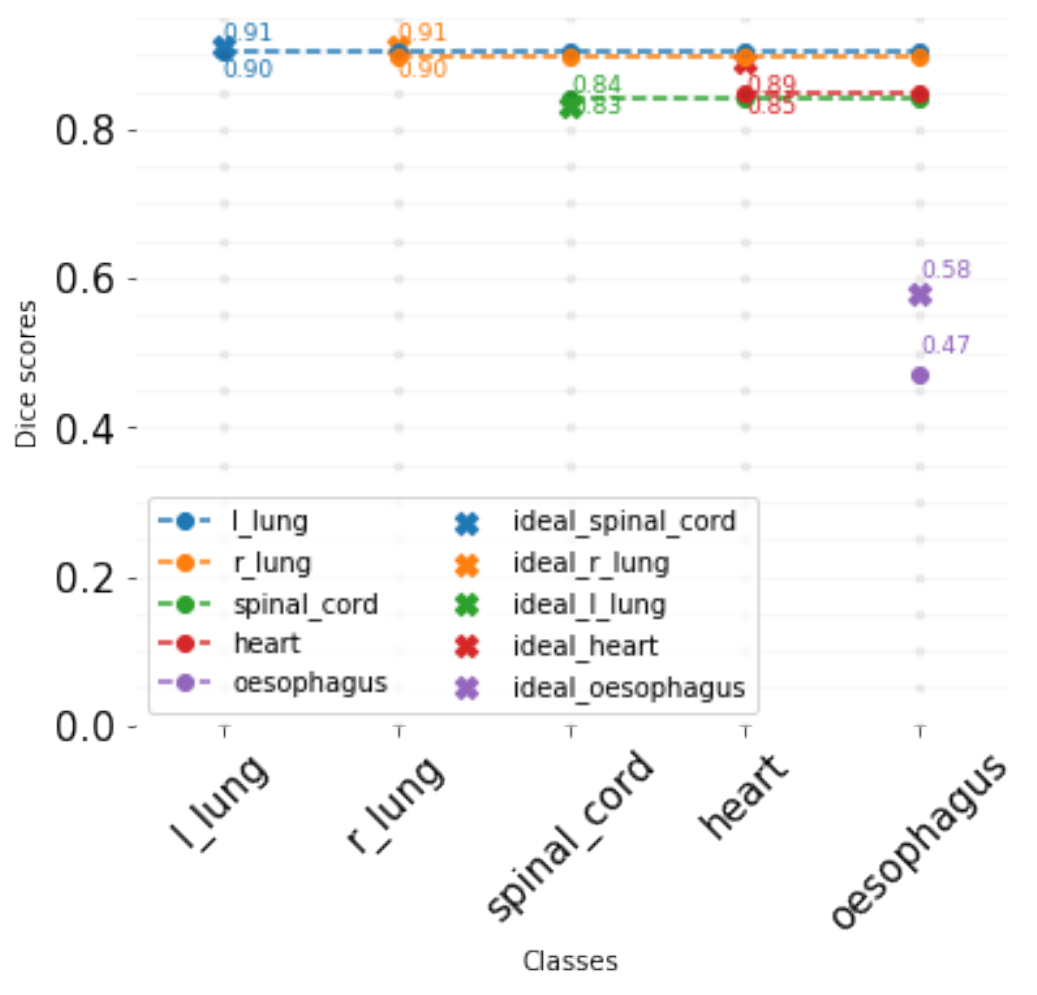} }}%
    \caption{Dice Scores for a) OrderB b) OrderC for ACLSeg}%
    \label{fig:order_b_c}%
\end{figure}
\end{center}

\begin{center}
\begin{figure}%
    \centering
    \subfloat[]{{\includegraphics[width=5cm]{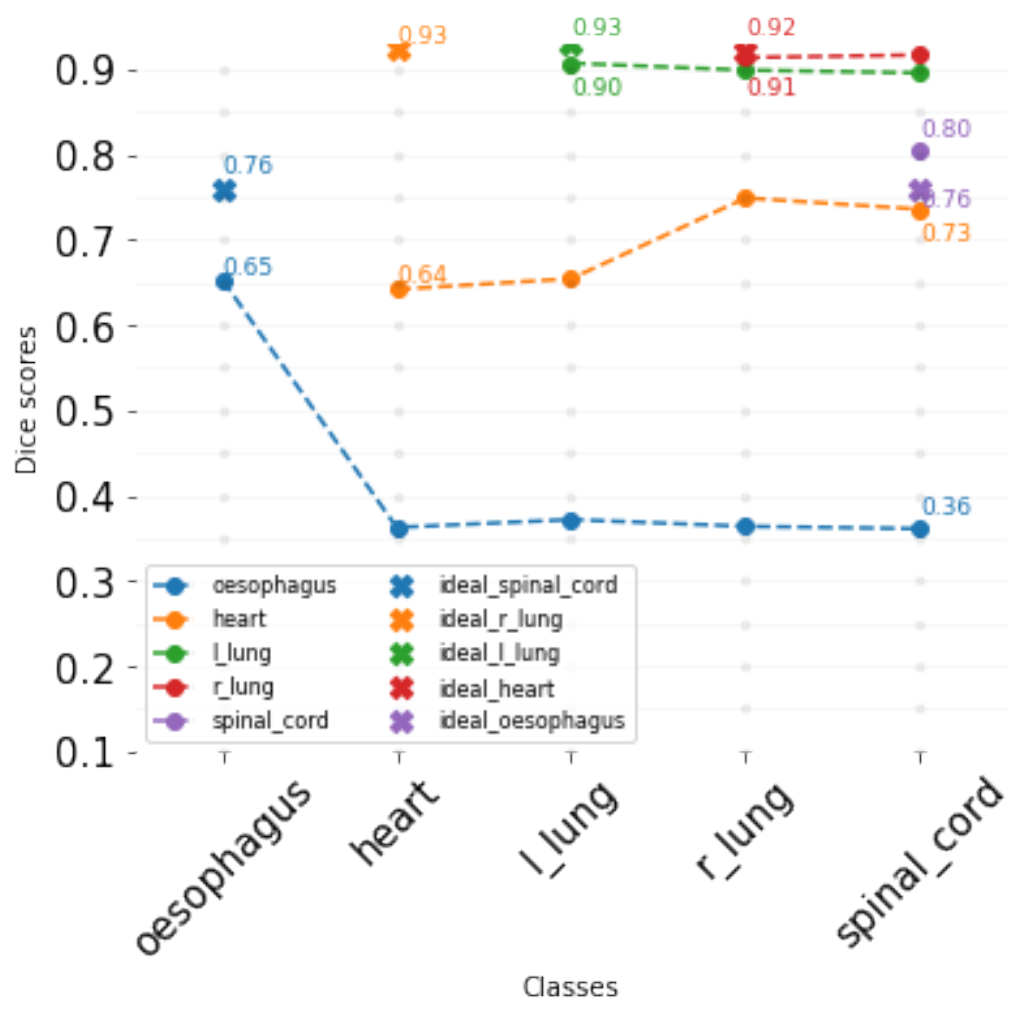} }}%
    \qquad
    \subfloat[]{{\includegraphics[width=5cm]{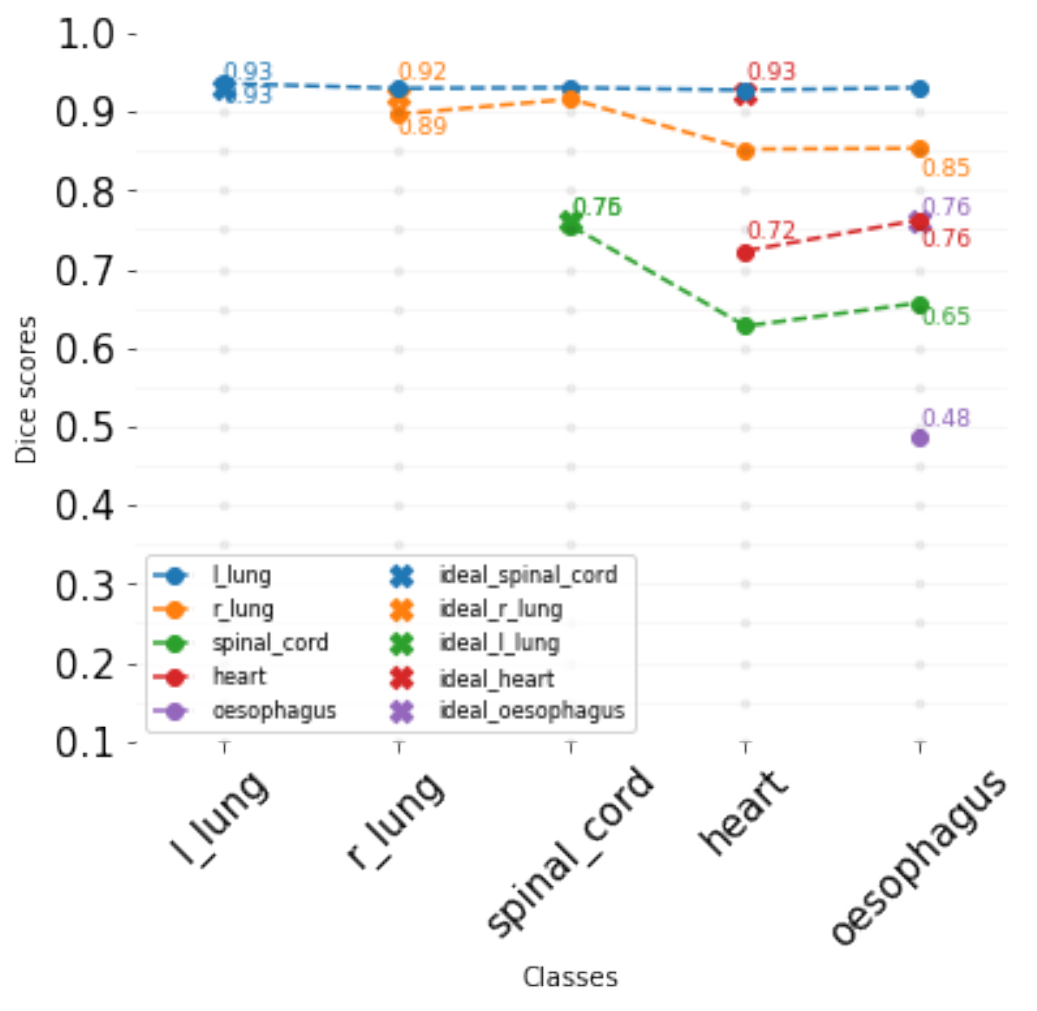} }}%
    \caption{Dice Scores for a) OrderB b) OrderC for LwF}%
    \label{fig:order_b_c_lwf}%
\end{figure}
\end{center}

\section{Dice scores of different task orders}
In Fig. \ref{fig:order_b_c}, and Fig. \ref{fig:order_b_c_lwf} we show dice scores for OrderB and OrderC for ACLSeg and LwF respectively. We observe that ACLSeg can maintain a consistent performance regardless of the order of the tasks. On the contrary, LwF exhibit degradation in performance of previously learned tasks to accommodate new ones. This degradation can be significant if the task is complex or different as with the Oesophagus in OrderB case, or can be slight as in OrderC case. Table \ref{tab_avg_dice} shows the overall mean dice scores, of the final model, achieved by both ACLSeg and LwF for different task orders. While ACLSeg shows comparable dice scores for different task orders, we deduct from Fig. \ref{fig:order_b_c} that the low  $\omega_{base}$ value in OrderB is not ascribed to forgetting the base class but to the model's inability to learn this class due to its difficulty. On the contrary, LwF behaviour, with the addition of new tasks, varies according to the order at which tasks are learned, even though it still shows comparable overall dice scores on the final model. 


\end{document}